\documentclass[11pt]{article}
\usepackage[no-math]{fontspec}
\usepackage[a4paper,margin=2.4cm]{geometry}
\usepackage{amsmath,amssymb}
\usepackage{booktabs,array,tabularx}
\usepackage{tikz,pgfplots}
\usetikzlibrary{arrows.meta,positioning,calc}
\pgfplotsset{compat=1.18}
\usepackage{flafter,placeins}
\usepackage[hidelinks]{hyperref}
\usepackage{url}
\newcommand{\Ax}{A_x}
\newcommand{\Ay}{A_y}
\newcommand{\best}[1]{\textbf{#1}}
\hypersetup{pdfauthor={Long Wang},pdftitle={From Task Supervision to Natural-Language Rules through Global Explanation Spaces}}
\title{From Task Supervision to Natural-Language Rules\\through Global Explanation Spaces}
\author{Long Wang\\
  {\small\href{mailto:alueas@126.com}{\texttt{alueas@126.com}}}\\
  {\small\href{https://orcid.org/0009-0006-8371-1562}{ORCID: 0009-0006-8371-1562}}}
\date{}
\begin{document}
\maketitle
\begin{abstract}
A customer-service reply depends on what the customer wants and what has already happened in the conversation. We learn global explanation spaces that describe dialogue states and response attributes using named dimensions shared across examples. Feedback from a supervised intent classifier guides revisions to the input dimensions. A mapping learned from demonstrated replies predicts which response attributes to emphasize given the dialogue state. On six ABCD customer-service intents, the revised features achieved a macro-F1 of 0.8615, compared with 0.6593 for TF-IDF. In a separate matching analysis, a nonlinear mapping improved mean reciprocal rank over ranking by a constant response profile. To guide generation, we approximated the original ridge mapping with a regression tree and translated its seven root-to-leaf paths into conditional rules. Under shared factual constraints, rule-guided generation succeeded on 29 of 30 dialogues, compared with 28 for the basic prompt, as judged by roles of the same language model. This exploratory study shows how to use a learned relationship between dialogue states and response priorities without running a separate feature extractor or numerical mapping for each reply.

\end{abstract}

\section{Introduction}
When a customer requests a refund, the appropriate reply depends in part on whether identity verification is complete. An intent label can distinguish a new request from a status inquiry. The recorded dialogue and reply can also show how the verification state affects what the agent should ask next. How can a language model use both kinds of supervision to learn distinctions that remain useful when it generates a reply?

Concept bottleneck models use human-readable concepts as intermediate features for prediction \cite{cbm}. Language-guided and annotation-free methods help specify and supervise those concepts \cite{labo,posthoc,labelfree}, and feedback or Bayesian search can improve the concept inventory \cite{escher,cocobm,bcllm}. These approaches motivate using task supervision to learn which distinctions the language model should extract from a dialogue.

We use a second set of dimensions to describe replies, including how much attention a reply gives to identity and order verification. Learning a mapping between the two spaces connects an input condition, such as incomplete verification, to a response priority, such as requesting the missing information. We test the mapping by ranking candidate replies, and we use it to guide generation either by producing instructions for each dialogue or by compiling a fixed set of conditional rules.

We call this approach \emph{global explanation space learning}. A space is global because every example uses the same dimension names, definitions, and scoring anchors, with values assigned from the individual dialogue or reply. Task-prediction feedback revises the input space $\Ax$. The mapping from $\Ax$ to the output space $\Ay$ then uses those features to predict response priorities. The compiled rules express the resulting conditions and priorities in natural language, so the generator can apply them directly to a dialogue.

The architecture draws on language as a representation for learned task structure \cite{latent}, text-based updates driven by task feedback \cite{protegi,bpo,opro,facet}, and symbolic approximation of trained predictors \cite{trepan}. Its contribution is to learn an explicit relationship between dialogue and reply attributes that can be examined through matching and used in generation.

The experiments use six customer-service intents from ABCD \cite{abcd}. Candidate matching tests the predicted response attributes before the language model has to act on them. The generation study follows the original ridge mapping through compilation and explores how different ways of expressing its priorities affect replies. Early rule formulations performed worse than the basic prompt, and their results help explain why the final rule set is evaluated under shared factual constraints.

\section{Related Work}
\subsection{Concept representations and language-guided space learning}
Concept bottleneck models predict labels through interpretable intermediate variables and allow interventions on those variables \cite{cbm}. Post-hoc concept bottlenecks transfer concepts into pretrained representations and support global model edits \cite{posthoc}; label-free concept bottlenecks reduce reliance on concept annotations \cite{labelfree}. LaBo uses language-generated candidate concepts and selects a discriminative, diverse bottleneck \cite{labo}. These approaches establish the predictive value of readable semantic coordinates and provide the foundation for our input explanation space.

The concept inventory itself can also be learned. Text Bottleneck Models discover and measure concepts for supervised text prediction \cite{tbm}. ESCHER improves a visual concept library using vision--language critics and performance history \cite{escher}. Conditional concept bottleneck models use agents to adjust a concept bank and refine concept scoring \cite{cocobm}. BC-LLM uses a language model as an extraction mechanism and prior within Bayesian concept search \cite{bcllm}. Our feedback stage shares the premise that a semantic space is revisable. We extend the learning object to include an output attribute space and its relation to the input space, then use the learned relation to specify generation behavior. Concept revision alone is therefore one component of the architecture rather than its complete contribution.

\subsection{Task feedback and structured prompt optimization}
ProTeGi converts task errors into textual gradients and searches over revised prompts \cite{protegi}. BPO learns prompt improvements from preference information \cite{bpo}, while OPRO uses previously evaluated solutions to guide language-model optimization \cite{opro}. Task Facet Learning structures prompts into semantic sections that can be added, edited, or deleted through feedback \cite{facet}. These methods already show that task evidence can improve black-box language-model behavior through text.

Here, prediction errors guide revisions to the definitions and scoring anchors used to measure each dialogue. The revised features also serve as inputs to a model that learns from demonstrated replies. This gives task feedback a role in the features used to learn response priorities.

\subsection{Attribute-controlled generation and executable rules}
Concept-based control also extends to generative models. CB-LLMs introduce interpretable concept representations for text classification and generation, including concept-based steering \cite{cbllm}. Post-hoc generative concept bottlenecks support interpretable control of pretrained image generators \cite{gencbm}. Thus, concept-guided generation itself is an established direction. The question here is how a supervised input--output relation over separately defined task and response spaces can be compiled into language that a generator executes.

Plug and Play Language Models guide a pretrained generator using attribute classifiers \cite{pplm}, and DExperts combines expert and anti-expert language models during decoding \cite{dexperts}. These methods establish attribute-based generation control. Our online path instead presents predicted attributes as natural-language instructions. The compiled path expresses the learned relation as conditional language rules, allowing the generator to interpret conditions directly from the dialogue.

Rule extraction also has a long history: TREPAN approximates a trained network by a decision tree \cite{trepan}. We use a regression-tree approximation of a learned output-attribute mapping and translate its conditions and leaf predictions into language. This links symbolic approximation to a language-model execution interface. It differs from using a tree solely as an explanation of a classifier: here the extracted object participates in producing the response. Learning with latent language provides a broader precedent for natural language as a task parameterization \cite{latent}. The present study connects that perspective to explicit input and output coordinates, supervised feedback, and compiled deployment.

\section{Global Explanation Space Learning}
\subsection{Shared semantic coordinates}
Let $x$ be a task input, $c$ its supervised task label, and $y$ a demonstrated output. The input explanation space at revision $t$ is
\begin{equation}
\Ax^{(t)}=\{(a_j,d_j,h_j)\}_{j=1}^{p_t},\qquad
z_x^{(t)}=\phi(x;\Ax^{(t)}).
\end{equation}
Here $a_j$ names a dimension, $d_j$ defines it operationally, and $h_j$ specifies scoring anchors and evidence requirements. A language model proposes the space and extracts its values. In this implementation, scores are integers from 0 to 4, accompanied by textual evidence. Shared definitions make feature values comparable across examples. They do not establish that a dimension is a causal variable or that its score is a calibrated probability.

The output space is
\begin{equation}
\Ay=\{(b_k,e_k,h'_k)\}_{k=1}^{q},\qquad
z_y=\psi(y;\Ay).
\end{equation}
The two spaces have different roles. An input dimension might measure identity-verification completeness; an output dimension might measure attention to identity and order verification. Their values need not agree. The mapping between them is learned from demonstrations.

A lightweight task predictor $g_\theta(z_x)$ learns the domain labels. Its errors provide feedback about which distinctions the input space captures. The current output extractor reads the response text alone. Attributes whose correctness depends on the dialogue can instead be defined through $\psi(x,y;\Ay)$; in the present study, actual response correctness is evaluated separately using context. Thus, an extracted attribute is a response characteristic, while its appropriateness is a task-conditioned question.

\subsection{Supervised feedback revises the input space}
The language model proposes an initial space from the task description and training examples. A supervised predictor then supplies errors, confusion patterns, permutation-based feature utility, and score distributions. The language model uses this feedback to retain, merge, remove, add, or redefine dimensions. Features are extracted again for each candidate, and candidates are evaluated under the same development protocol.

Writing the development loss as $V(\Ax)$, the implemented acceptance rule is
\begin{equation}
V(\widetilde\Ax)\leq V(\Ax)-\epsilon,\qquad \epsilon=0.01.
\label{eq:accept}
\end{equation}
The experiment contains one revision round. A textual-feedback candidate uses error examples and labels; a numerical-feedback candidate additionally receives confusion, utility, and score summaries. When both are accepted, the implemented protocol gives priority to the numerical candidate. Acceptance records the observed space update; it does not guarantee lower test risk. Under a fixed evaluation, Equation~\ref{eq:accept} makes accepted development losses nonincreasing by construction.

\subsection{Learning the input--output relation}
Given demonstrations $\mathcal D_{xy}$, extracted pairs $(z_{x,i},z_{y,i})$ train a mapping
\begin{equation}
F_\eta:z_x\longmapsto\widehat z_y,\qquad
\mathcal L_{\mathrm{map}}(\eta)=\frac{1}{n}\sum_{i=1}^{n}
\|F_\eta(z_{x,i})-z_{y,i}\|_2^2.
\label{eq:map}
\end{equation}
The mapping predicts the attribute profile of an associated response. Matching a candidate response to the predicted profile uses
\begin{equation}
s(x,y)=-\|F_\eta(\phi(x))-\psi(y)\|_2^2.
\label{eq:matching}
\end{equation}
Candidate ranking makes the relation observable without conflating it with fluent generation. Matching can also help select demonstrations for later relation learning; that extension belongs to the architecture, while the current implementation learns from the recorded reference replies.

The primary mapping is standardized ridge regression, with predictions clipped to $[0,4]$. Alternative linear and nonlinear mappings examine how the statistical model affects use of the same coordinates. The squared-distance objective treats the anchored ordinal scores as numerical coordinates with equal weight. Other metrics or calibrated scales are possible implementations of the relation-learning component.

\subsection{From learned attributes to language execution}
In online generation, the mapping predicts an output profile for each input. A compiler $\Pi$ turns dimension definitions and predicted priorities into instructions:
\begin{equation}
\widehat z_y=F_\eta(\phi(x)),\qquad
p_x=\Pi(\Ax,\Ay,\widehat z_y),\qquad
\widehat y=G(x,p_x).
\end{equation}
The generator $G$ retains its pretrained language abilities. Attribute strength specifies what to address and with what emphasis. Factual constraints govern what information is available to say. For example, emphasizing timeline accuracy should require grounded statements about timing rather than encourage an invented processing duration.

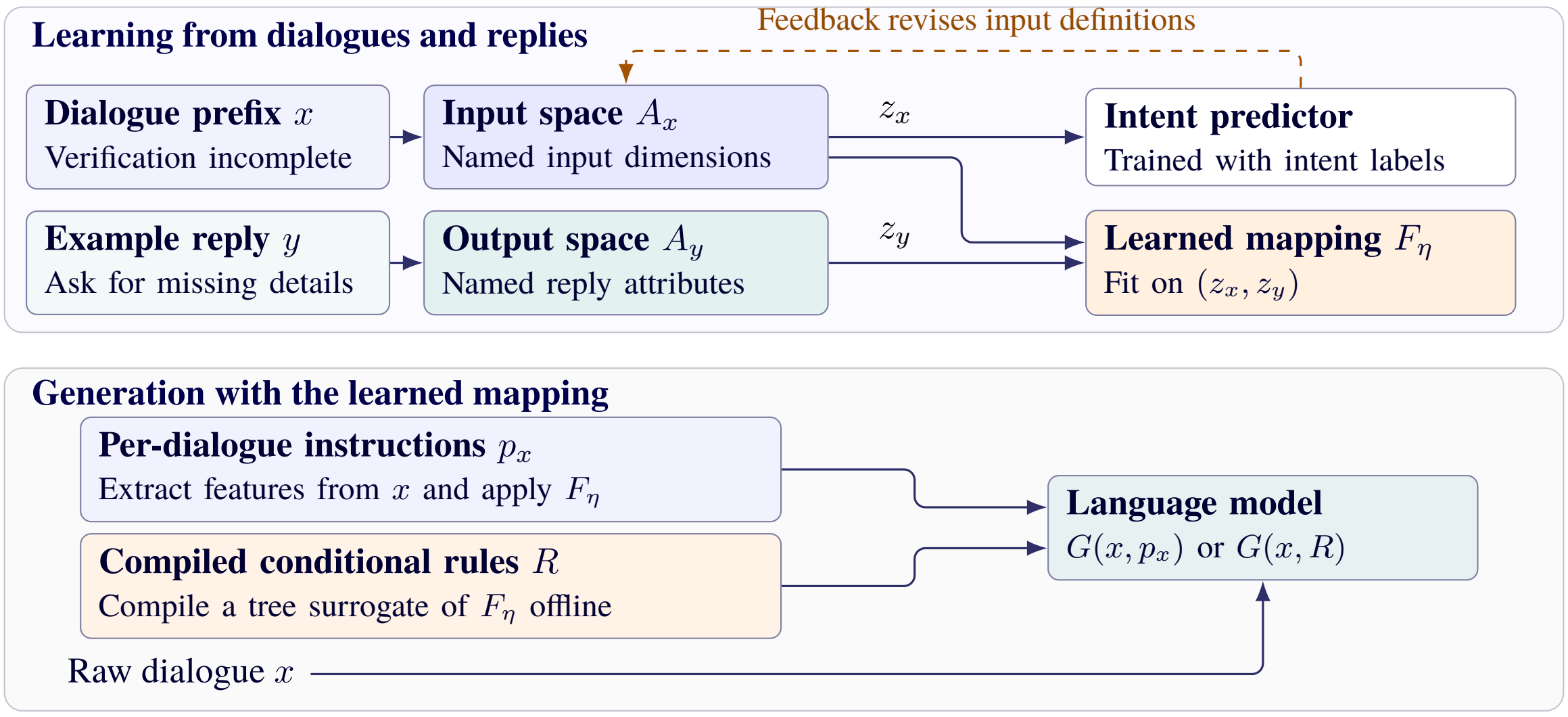
\begin{figure}[tbp]
\centering
\begin{tikzpicture}[>=Latex,font=\small,
 box/.style={rounded corners=3pt,inner xsep=5pt,inner ysep=4pt,align=left,draw=blue!28!black!50,text=blue!20!black},
 scene/.style={box,text width=3.2cm},
 space/.style={box,text width=3.6cm},
 model/.style={box,text width=3.85cm},
 route/.style={box,text width=6.5cm},
 flow/.style={->,semithick,draw=blue!28!black!82,rounded corners=3pt},
 feedback/.style={flow,dashed,draw=orange!64!black},
 heading/.style={anchor=west,inner sep=0pt,font=\small\bfseries,text=blue!30!black}]
\draw[rounded corners=5pt,fill=blue!2,draw=blue!18!black!23] (0,1.0) rectangle (15.25,4.18);
\node[heading] at (.26,3.88) {Learning from dialogues and replies};
\node[scene,fill=blue!5] (dialogue) at (1.99,2.91) {\textbf{Dialogue prefix $x$}\\ {\footnotesize Verification incomplete}};
\node[scene,fill=teal!5] (reply) at (1.99,1.68) {\textbf{Example reply $y$}\\ {\footnotesize Ask for missing details}};
\node[space,fill=blue!9] (input) at (6.08,2.91) {\textbf{Input space $\Ax$}\\ {\footnotesize Named input dimensions}};
\node[space,fill=teal!11] (output) at (6.08,1.68) {\textbf{Output space $\Ay$}\\ {\footnotesize Named reply attributes}};
\node[model,fill=white] (predictor) at (12.68,2.91) {\textbf{Intent predictor}\\ {\footnotesize Trained with intent labels}};
\node[model,fill=orange!12] (mapping) at (12.68,1.68) {\textbf{Learned mapping $F_\eta$}\\ {\footnotesize Fit on $(z_x,z_y)$}};
\draw[flow] (dialogue) -- (input);
\draw[flow] (reply) -- (output);
\draw[flow] (input.east) -- node[above,pos=.26] {$z_x$} (predictor.west);
\draw[flow] ([yshift=-2mm]input.east) -- ++(1.30,0) |- ([yshift=2mm]mapping.west);
\draw[flow] (output.east) -- node[above,pos=.26] {$z_y$} (mapping.west);
\draw[feedback] (predictor.north) -- ++(0,.36) -| node[pos=.24,above,font=\footnotesize,text=orange!58!black] {Feedback revises input definitions} (input.north);

\draw[rounded corners=5pt,fill=black!2,draw=blue!18!black!23] (0,-2.70) rectangle (15.25,.65);
\node[heading] at (.26,.38) {Generation with the learned mapping};
\node[route,fill=blue!5] (online) at (4.17,-.34) {\textbf{Per-dialogue instructions $p_x$}\\ {\footnotesize Extract features from $x$ and apply $F_\eta$}};
\node[route,fill=orange!10] (rules) at (4.17,-1.48) {\textbf{Compiled conditional rules $R$}\\ {\footnotesize Compile a tree surrogate of $F_\eta$ offline}};
\node[model,fill=teal!10] (gen) at (12.31,-.91) {\textbf{Language model}\\ {\footnotesize $G(x,p_x)$ or $G(x,R)$}};
\node[anchor=west,inner sep=0pt,text=blue!20!black] (raw) at (.61,-2.34) {Raw dialogue $x$};
\draw[flow] (online.east) -- ++(1.30,0) |- ([yshift=2mm]gen.west);
\draw[flow] (rules.east) -- ++(1.30,0) |- ([yshift=-2mm]gen.west);
\draw[flow] ([xshift=1.5mm]raw.east) -- (12.31,-2.34) -- (gen.south);
\end{tikzpicture}
\caption{Learning and generation through explanation spaces, illustrated with a refund dialogue. The intent predictor sends feedback to the input dimensions. Dialogue and demonstrated-reply features train the mapping. At inference, the language model receives the raw dialogue and either instructions computed for that dialogue or a set of rules compiled offline.}
\label{fig:framework}
\end{figure}

For compiled-rule generation, a small regression tree $T$ approximates the mapping on training inputs:
\begin{equation}
\min_T\sum_{i=1}^{n}\|T(z_{x,i})-F_\eta(z_{x,i})\|_2^2.
\end{equation}
Each root-to-leaf path becomes a conditional language rule, and leaf predictions become output-attribute priorities. The named conditions and response priorities form a global instruction set $R$. Figure~\ref{fig:framework} summarizes the architecture. At deployment,
\begin{equation}
\widehat y=G(x,R).
\label{eq:compiled}
\end{equation}
The language model interprets the dialogue, determines applicable conditions, and executes the instructions. There is no separate online call to $\phi$, $F_\eta$, or $T$. This removes those numerical components from deployment while retaining the language model.

\subsection{Statistical interpretation of relation learning}
Squared-error relation learning has a precise population target. For random extracted coordinates $Z_x,Z_y$ with $\mathbb E\|Z_y\|_2^2<\infty$, let $m(Z_x)=\mathbb E[Z_y\mid Z_x]$. Conditional-expectation orthogonality gives
\begin{align}
\mathbb E\|Z_y-F(Z_x)\|_2^2
&=\mathbb E\|Z_y-m(Z_x)\|_2^2\nonumber\\
&\quad+\mathbb E\|m(Z_x)-F(Z_x)\|_2^2.
\label{eq:risk}
\end{align}
The optimal measurable mapping is therefore $F^\star=m$. Relative to the constant $\mu=\mathbb E[Z_y]$, its risk reduction is
\begin{equation}
\mathbb E\|m(Z_x)-\mu\|_2^2\geq0.
\end{equation}
This standard projection identity explains what the mapping can learn: input-dependent variation in response attributes. It provides no finite-sample guarantee for a chosen model family and no automatic guarantee for candidate rank or generated text. Those outcomes depend on estimation, attribute coverage, compilation, and execution. We evaluate them directly.

\section{Experimental Design}
\subsection{Data, information access, and model settings}
We use the product-defect workflow of ABCD v1.1 \cite{abcd}, covering refund initiation, refund modification, refund status, and returns due to stains, color, or size. Within the official splits, dialogue IDs are ordered by SHA-256 and subsampled by intent: 10 training, 5 development, and 9 test dialogues per intent, yielding 60/30/54 examples. The input contains the dialogue prefix and recorded actions preceding a target agent reply. The target reply is excluded from input feature extraction.

Each test example has 15 response candidates: its recorded reference and 14 deterministically selected alternatives. All matching methods use the same candidate lists. Ties are resolved by stable sorting, with the reference stored at index zero. This ordering can favor the reference when distances tie. A diagnostic that uses the reference's own output vector as the prediction is a self-distance check, not independent evidence that the output space identifies appropriate replies. Appendix~\ref{app:ranking} explains the distinction.

Space proposal and revision use task supervision. The input extractor receives only the dialogue prefix and sample identifier, together with the shared space definition. Demonstration replies supply output attributes for mapping training. All language-model roles use Doubao Seed 2.1 Pro at temperature zero, with thinking disabled. The model identifier and role-specific prompts are specified in Appendix~\ref{app:implementation} and the accompanying reproducibility materials. The initial input proposal contains 12 dimensions; review retains 8; revision produces 10. The output space contains 10 dimensions. All have five scoring levels.

\subsection{Learning and comparison protocols}
The task predictor is standardized, class-weighted logistic regression. Five-fold cross-validation on training examples selects $C=10$ from $\{0.01,0.1,1,10\}$. TF-IDF with logistic regression is a lexical supervised reference. It does not share the language-model semantic prior used by the explanation features.

For relation learning, five-fold training cross-validation selects ridge regularization $\alpha=100$ from $\{0.01,0.1,1,10,100\}$ by output-coordinate mean squared error. The nonlinear exploration includes a random forest with 200 trees, unrestricted depth, and random state 42. We report MRR@15, Hit@1, and Hit@3 on 54 test examples. Subsequent model comparisons and training-prefix curves use already observed data and are reported as exploratory analyses. The curve uses the first $n$ rows of the same training array, changing both sample count and prefix composition.

The compiled generator uses the original ridge mapping with $\alpha=100$. Its surrogate is a regression tree of maximum depth 3, minimum leaf size 5, and random state 0. It approximates mapping predictions on 60 training inputs. The nonlinear matching exploration was conducted separately and was not connected to this generation pipeline.

\subsection{Generation evaluation}
Generation evaluation uses five test examples per intent, for 30 examples in total. Task success is the conjunction of three binary conditions: the next step is correct, the response follows the applicable policy, and it contains no unsupported factual claims. Helpfulness and naturalness are scored separately. Evaluation uses two prompt-defined roles of the same language model, with arbitration of disagreements. This is model-based evaluation rather than human annotation or independent cross-model assessment. Prior work documents both the usefulness and potential biases of language-model judges \cite{geval,mtbench}; here the role structure does not remove the dependence created by using one model throughout.

The generation study progressively revises how numerical attributes are expressed as instructions. It compares the basic prompt, online mapping instructions, and compiled rules under each version's recorded protocol. The main table reports the version with shared constraints against unsupported facts. Other versions, reference-attribute diagnostics, and direct prompting comparisons remain in Appendix~\ref{app:interfaces}. Some outputs are reused and jointly reevaluated between versions, so a score change across versions need not imply a newly generated response.

\begin{table}[tbp]
\centering\small
\caption{Evaluation objects along the supervision pathway. Each result addresses the indicated component rather than substituting for an end-to-end quality measure.}
\label{tab:objects}
\begin{tabularx}{\linewidth}{p{2.8cm}XX}
\toprule
Component & Operation & Observable outcome\\
\midrule
Input space & Train a task classifier on $z_x$ & Test Macro-F1, accuracy, cross-entropy\\
Space revision & Re-extract features after feedback & Candidate development loss and space changes\\
Output relation & Fit $z_x\mapsto z_y$ & Candidate rank and training-prefix behavior\\
Generation & Approximate and compile the mapping & Surrogate fidelity and response success\\
\bottomrule
\end{tabularx}
\end{table}

\section{Results}
\subsection{Dialogue features improve intent classification}
The revised input space yields test Macro-F1 0.8615 and accuracy 0.8704, compared with 0.6593 and 0.6667 for TF-IDF (Table~\ref{tab:task}). The Macro-F1 difference is 0.2022. Cross-entropy is also lower. These results show that the shared semantic dimensions provide useful supervised features for the six-intent task. The comparison measures the resulting feature pipelines, including the explanation extractor's pretrained semantic knowledge.

\begin{table}[tbp]
\centering\small
\caption{Task prediction on 54 test dialogues. Higher Macro-F1 and accuracy and lower balanced cross-entropy (CE) are better. Bold indicates the best comparable value in each column; it does not indicate statistical significance. The same convention applies to subsequent comparison tables.}
\label{tab:task}
\begin{tabular}{lccc}
\toprule
Representation and predictor & Macro-F1 $\uparrow$ & Accuracy $\uparrow$ & CE $\downarrow$\\
\midrule
Explanation space + logistic regression & \best{0.8615} & \best{0.8704} & \best{0.8319}\\
TF-IDF + logistic regression & 0.6593 & 0.6667 & 1.5096\\
Majority class & --- & 0.1667 & ---\\
\bottomrule
\end{tabular}
\end{table}

\subsection{Feedback revises how dialogue features are scored}
One numerical-feedback revision reduces candidate-stage development cross-entropy from 1.6309 to 0.7616, a reduction of 0.8693 or 53.30\%. The textual-feedback candidate achieves 0.7236 (Table~\ref{tab:revision}). Both candidates meet the acceptance rule; numerical-feedback priority selects the numerical candidate even though the textual candidate has the lower loss. Thus, the observations establish that feedback can produce useful space revisions, while providing no evidence that the extra numerical summaries outperform textual feedback in this round.

\begin{table}[tbp]
\centering\small
\caption{One-round candidate evaluation on the development split. Dimension count is descriptive. The final selected-space re-extraction has development CE 0.6288 and is a separate extraction stage.}
\label{tab:revision}
\begin{tabular}{lccc}
\toprule
Space & Dimensions & CE $\downarrow$ & Macro-F1 $\uparrow$\\
\midrule
Reviewed initial space & 8 & 1.6309 & ---\\
Textual-feedback candidate & 10 & \best{0.7236} & \best{0.8008}\\
Numerical-feedback candidate & 10 & 0.7616 & 0.6245\\
\bottomrule
\end{tabular}
\end{table}

The revision introduces or reorganizes dimensions for refund lifecycle, requested service action, product-attribute evidence, and observability of intent evidence. These definitions turn distinctions present in task errors into scoring criteria shared across examples. Space revision therefore changes the semantic representation supplied to the task predictor, not just its fitted coefficients. Candidate evaluation and final re-extraction are reported separately because they are different recorded stages.

\subsection{Predicted response attributes help rank candidate replies}
The random-forest mapping obtains MRR@15 0.3345, Hit@1 0.1481, and Hit@3 0.3704 (Table~\ref{tab:mapping}). The training-output mean obtains MRR 0.2881. The cross-validated ridge mapping reaches 0.2838; weaker regularization with $\alpha=1$ reaches 0.3265 and gives the highest Hit@1 in this comparison. These differences show that model choice affects how the coordinates support candidate matching. In particular, the mapping selected by coordinate mean squared error need not be best on a ranking metric.

\begin{table}[tbp]
\centering\small
\caption{Output-relation matching on 54 examples with 15 candidates each. All metrics are higher-is-better. The nonlinear and alternative-regularization comparisons are exploratory.}
\label{tab:mapping}
\begin{tabular}{lccc}
\toprule
Relation model & MRR@15 & Hit@1 & Hit@3\\
\midrule
Uniform random ranking (expectation) & 0.2212 & 0.0667 & 0.2000\\
Training-output mean & 0.2881 & 0.1111 & 0.3148\\
Ridge, $\alpha=100$ & 0.2838 & 0.0926 & 0.3519\\
Ridge, $\alpha=1$ & 0.3265 & \best{0.1667} & 0.3519\\
Random forest & \best{0.3345} & 0.1481 & \best{0.3704}\\
\bottomrule
\end{tabular}
\end{table}

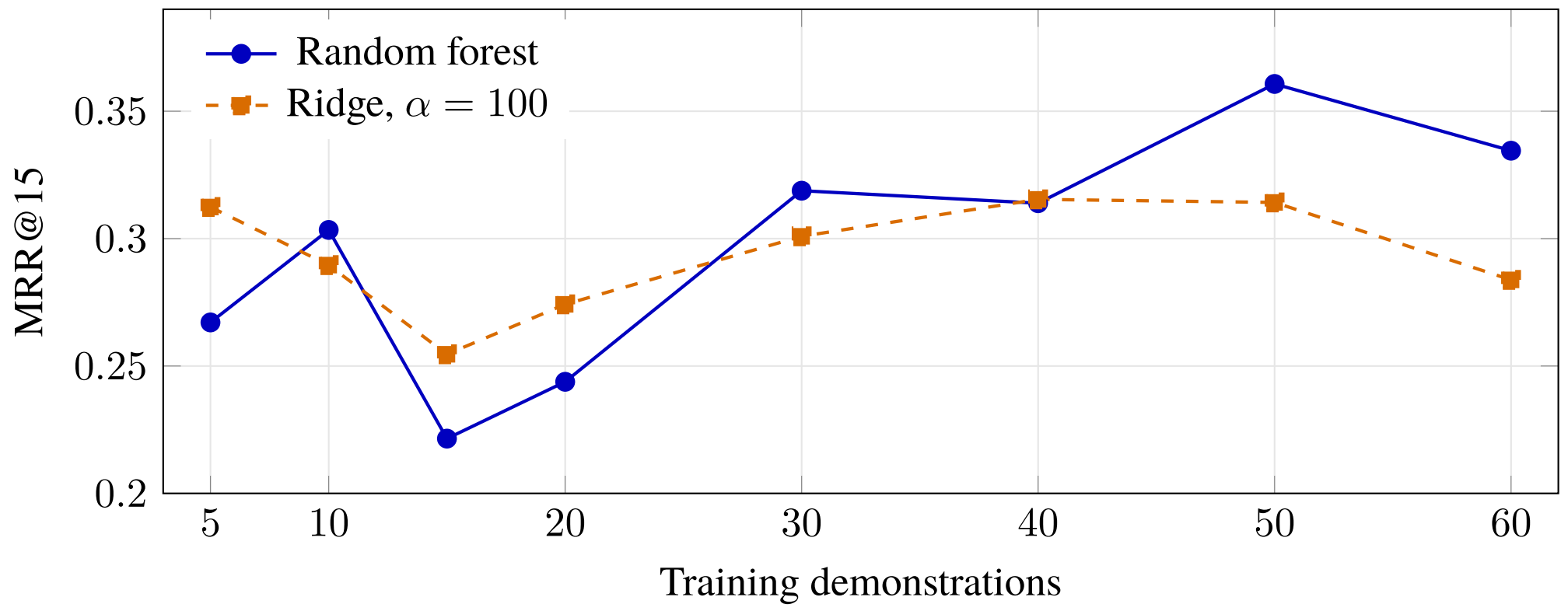
\begin{figure}[tbp]
\centering
\begin{tikzpicture}
\begin{axis}[width=.83\linewidth,height=5.7cm,xlabel={Training demonstrations},ylabel={MRR@15},
 xmin=3,xmax=62,ymin=.20,ymax=.39,xtick={5,10,20,30,40,50,60},
 ytick={.20,.25,.30,.35},grid=major,grid style={black!10},
 legend style={at={(.02,.98)},anchor=north west,font=\small,draw=none},
 tick label style={font=\small},label style={font=\small}]
\addplot[blue!75!black,thick,mark=*] coordinates {(5,.2671)(10,.3034)(15,.2215)(20,.2438)(30,.3188)(40,.3139)(50,.3607)(60,.3345)};
\addlegendentry{Random forest}
\addplot[orange!85!black,thick,dashed,mark=square*] coordinates {(5,.3124)(10,.2896)(15,.2546)(20,.2742)(30,.3009)(40,.3154)(50,.3142)(60,.2838)};
\addlegendentry{Ridge, $\alpha=100$}
\end{axis}
\end{tikzpicture}
\caption{Exploratory relation-learning trajectories on the same test set. Each point uses a prefix of the original training array. Lines connect observations; they do not imply monotonic improvement. These are candidate-matching results, with no generation evaluation along the curve.}
\label{fig:curve}
\end{figure}

Training-prefix evaluation provides a second view of relation learning. Random-forest MRR rises from 0.2671 at 5 demonstrations to 0.3345 at 60, reaching 0.3607 at 50 (Figure~\ref{fig:curve}). A line fitted to all eight observations has slope 0.00177 per example and $R^2=0.5453$. The observed trend is positive and nonmonotonic. Prefix composition varies, and the curve has no repeated data draws or seed replication. It describes learning behavior in this exploratory sequence rather than a population scaling law. Complete values for both mappings appear in Appendix~\ref{app:models}.

\subsection{A learned relation can be compiled for generation}
The original ridge mapping is approximated by a depth-three tree with seven leaves, producing seven conditional language rules. Against the mapping predictions on the training inputs, the tree has $R^2=0.6869$ and mean squared error 0.00754. These are training-fidelity measures for the surrogate, distinct from fidelity to demonstrated output attributes and from the quality of generated responses.

Under the version with shared constraints against unsupported facts, compiled-rule generation succeeds on 29 of 30 examples, the basic prompt on 28, and online mapping on 25 (Table~\ref{tab:generation}). Compiled rules win on two cases that the basic prompt fails and lose on one case that the basic prompt passes, for a net difference of one response or 3.33 percentage points. Helpfulness is 4.12 versus 3.93; naturalness is 4.53 versus 4.57. These observations establish a functioning generation path from a learned mapping to executable rules. They do not establish statistically reliable superiority of compiled rules over the basic prompt.

\begin{table}[tbp]
\centering\small
\caption{Generation under shared factual constraints, evaluated on 30 dialogues by roles of the same language model. All three columns are higher-is-better. Numerical winners differ by outcome.}
\label{tab:generation}
\begin{tabular}{lccc}
\toprule
Generation method & Successful cases & Helpfulness & Naturalness\\
\midrule
Basic prompt & 28/30 & 3.93 & \best{4.57}\\
Online mapping & 25/30 & \best{4.18} & 4.38\\
Compiled rules & \best{29/30} & 4.12 & 4.53\\
\bottomrule
\end{tabular}
\end{table}

The original attribute instructions yield 18 successes for compiled rules and 25 for the basic prompt. A later version that limits attention to two attributes and requests shorter responses yields 28 and 29, respectively. The version with shared factual constraints was developed during this exploration. The sequence shows how much observed success can vary with the instructions that express the learned response priorities.

\section{Discussion}
\subsection{What dialogue features tell us about replies}
The revision makes distinctions such as refund stage and evidence availability explicit in the scoring criteria. Their value for classifying a dialogue and their value for predicting how to respond are separate questions. In the second case, the learned mapping predicts which attributes a reply should emphasize given the dialogue features. Its predictions can be evaluated by ranking candidate replies before any instructions are written for the generator.

The random forest improves matching over the constant-profile reference, while the generation experiment uses rules derived from ridge regression. Improving candidate rank, approximating a fitted mapping, and improving the generated reply are therefore separate tests. In particular, training fidelity measures how closely the tree reproduces the ridge predictions. The generation results also depend on how the resulting priorities are expressed to the language model.

In the current study, the dialogue prefixes and recorded replies supply the input features and response attributes for fitting the mapping. A further use of the same spaces would be to select demonstrations by task score or by how well their response attributes match the dialogue before fitting the mapping.

\subsection{What compilation transfers}
Compiled deployment transfers a fitted relation into a form the language model can interpret. The regression tree provides explicit conditions and output priorities, while the generator supplies the semantic interpretation needed to apply them to raw dialogue. Numerical surrogate fidelity therefore measures only one part of the transfer. A separate feature extraction call has been removed, but semantic inference still occurs inside the language model.

A rule can ask the model to address the timing of a refund. The dialogue still determines whether the model can state a processing duration. The early instructions sometimes encouraged unsupported specificity. The later versions clarify that the model should honor an attribute priority using the facts available in the dialogue.

\subsection{Scope and limitations}
The evidence comes from a small, balanced subset of one customer-service workflow and one fixed language model. Classification uses 54 test examples; generation uses 30. Space construction, nonlinear mapping comparisons, and interface revisions form an exploratory process. Test-set observations informed subsequent analyses, so the final generation version and prefix trend are not independent confirmations. Additional domains, models, independently chosen evaluation material, and variation in supervision would test how the mechanism generalizes.

The semantic dimensions are named and operationally defined, but extraction is model-dependent. Output scoring currently reads replies without their input context; qualities such as policy compliance require contextual interpretation, supplied separately in generation evaluation. Ordinal attributes, equal-weight distances, and the ridge mapping are implementation choices rather than guarantees of calibrated semantic measurement. The stronger BGE-based matching references in Appendix~\ref{app:bge} also show that readable coordinates do not imply the best available candidate-ranking performance.

Finally, generation is assessed by roles of the same model that constructs and executes the spaces. Role separation and arbitration are useful procedural distinctions, but they do not supply independent human judgments. Rule fidelity is measured on training inputs, and the small paired generation difference does not support a general superiority claim. These qualifications locate the current evidence: the supervision-to-execution pathway is realized, while its reliability and comparative advantages across conditions remain questions for further study.

\section{Conclusion}
Global explanation space learning makes the relationship between a dialogue state and its response priorities available to both supervised models and language generation. On ABCD, task errors helped revise the dialogue dimensions, and predictions in the reply space supplied an additional way to rank candidate responses. Compiling the ridge mapping let the generator apply learned conditions and priorities directly to the dialogue. By retaining the dimensions and the rules in a readable form, this architecture makes it possible to examine what the mapping learned and what the generation instructions asked the model to do with it.

\FloatBarrier
\clearpage
\appendix
\section{Additional Relation Models and Training Prefixes}
\label{app:models}
Table~\ref{tab:allmodels} compares all evaluated mapping specifications. These models use the same explanation coordinates and candidate lists. Their comparison was formed during exploration, rather than fixed before observing results. Table~\ref{tab:curve} supplies every point in Figure~\ref{fig:curve}. Prefix rows are in the original training-array order; changing $n$ can change intent coverage as well as sample size.

\begin{table}[htbp]
\centering\small
\caption{Complete mapping-model comparison on 54 test examples. All metrics are higher-is-better. Tied best values and reference winners are included in the emphasis convention.}
\label{tab:allmodels}
\begin{tabular}{lccc}
\toprule
Model & MRR@15 & Hit@1 & Hit@3\\
\midrule
Ridge, $\alpha=1$ & 0.3265 & \best{0.1667} & 0.3519\\
Ridge, $\alpha=10$ & 0.2992 & 0.1296 & 0.2963\\
Ridge, $\alpha=100$ & 0.2838 & 0.0926 & 0.3519\\
Quadratic features + ridge & 0.3162 & 0.1481 & 0.3333\\
Cubic features + ridge & 0.3010 & 0.1296 & 0.3519\\
Random forest & \best{0.3345} & 0.1481 & \best{0.3704}\\
Shallow random forest & 0.3187 & 0.1481 & 0.3148\\
Small multilayer perceptron & 0.2976 & 0.1481 & 0.2963\\
Medium multilayer perceptron & 0.3060 & 0.1296 & 0.3333\\
Nearest neighbors, $k=3$ & 0.3077 & 0.1481 & 0.2963\\
Nearest neighbors, $k=5$ & 0.3054 & 0.1296 & 0.3148\\
\bottomrule
\end{tabular}
\end{table}

\begin{table}[htbp]
\centering\small
\caption{Complete training-prefix observations, MRR@15. Each row compares the two models at the same $n$; bold marks the larger value in that row, not a preferred training-set size.}
\label{tab:curve}
\begin{tabular}{rcc@{\qquad}rcc}
\toprule
$n$ & Forest & Ridge & $n$ & Forest & Ridge\\
\midrule
5 & 0.2671 & \best{0.3124} & 30 & \best{0.3188} & 0.3009\\
10 & \best{0.3034} & 0.2896 & 40 & 0.3139 & \best{0.3154}\\
15 & 0.2215 & \best{0.2546} & 50 & \best{0.3607} & 0.3142\\
20 & 0.2438 & \best{0.2742} & 60 & \best{0.3345} & 0.2838\\
\bottomrule
\end{tabular}
\end{table}

\section{Generation Interfaces and Diagnostic Comparisons}
\label{app:interfaces}
All interface versions use the same 30 evaluation dialogues. Table~\ref{tab:interfaces} preserves their task-success counts. Reference-attribute generation receives the recorded reply's attribute vector; it is an information-advantaged diagnostic rather than a deployable method. The original three-example retrieval prompt succeeds on 22/30 dialogues and direct prompt optimization on 17/30. The original reference-attribute outputs are evaluated separately from the original formal-method comparison.

\begin{table}[htbp]
\centering\small
\caption{Complete interface trajectory. Counts are out of 30; bold marks within-row numerical winners, including ties and the information-advantaged diagnostic. Cross-row changes include prompt and evaluation changes.}
\label{tab:interfaces}
\begin{tabularx}{\linewidth}{Xcccc}
\toprule
Interface version & Basic & Online & Rules & Reference\\
\midrule
Original attribute instructions & 25 & 12 & 18 & $\best{26}^{\dagger}$\\
Factual guard for mapping/reference only & --- & \best{29} & --- & 28\\
Shared factual guard & 28 & 25 & \best{29} & 25\\
Two selected attributes, short response & \best{29} & 25 & 28 & 24\\
\bottomrule
\end{tabularx}
\par\smallskip
\begin{minipage}{.97\linewidth}\footnotesize
$\dagger$ Separately evaluated reference-attribute diagnostic. ``Reference'' denotes access to demonstrated output attributes during generation, not reference self-distance in candidate ranking. The shared-guard stage reuses earlier online/reference outputs and reevaluates them jointly. Consequently, score changes may reflect evaluation presentation as well as generation changes. Shared factual constraints do not mean that all prompt text is identical.
\end{minipage}
\end{table}

The original generation archive contains six methods and 54 attempted outputs per method. One retrieval-prompt request fails; the archive retains the error. The reported generation evaluation uses the fixed 30-dialogue subset, with the success definition held explicit at each stage. Full texts, role judgments, and version-specific prompts are retained in the experiment records.

\section{Embedding-Based Matching References}
\label{app:bge}
Table~\ref{tab:bge} places explanation-coordinate matching alongside BGE-based references \cite{bge}. Direct matching uses cosine similarity between input and candidate embeddings. The nearest-neighbor method averages output attributes from five neighboring training inputs; the regression method maps input embeddings to output attributes. BGE uses the last four turns, while explanation extraction reads the full dialogue prefix. Thus these are performance references with different context and model specifications, rather than an ablation isolating representation readability.

\begin{table}[htbp]
\centering\small
\caption{BGE-based candidate-matching references. All metrics are higher-is-better. The explanation-space forest is repeated to make the comparison visible.}
\label{tab:bge}
\begin{tabular}{lccc}
\toprule
Method & MRR@15 & Hit@1 & Hit@3\\
\midrule
Explanation-space random forest & 0.3345 & 0.1481 & 0.3704\\
Direct BGE cosine & 0.2892 & 0.1111 & 0.3148\\
BGE, five-neighbor attribute mean & \best{0.3809} & 0.2037 & \best{0.4630}\\
BGE-to-attribute ridge regression & 0.3736 & \best{0.2222} & 0.3704\\
\bottomrule
\end{tabular}
\end{table}

\section{Ranking Interpretation and Coordinate Definitions}
\label{app:ranking}
The mean reciprocal rank for a uniform random ordering of 15 candidates is $H_{15}/15=0.2212$, where $H_{15}=\sum_{r=1}^{15}1/r$. Hit@1 and Hit@3 expectations are $1/15$ and $3/15$. These reference values describe uniformly randomized ranks. Deterministic tie resolution in the implemented matching function need not have those expectations.

If a prediction is set equal to the reference response's own attribute vector, its squared self-distance is zero. Stable reference-first ordering then gives rank one even when other candidates tie. In the stored diagnostic, two examples have tied nearest candidates; an all-zero-vector control also gives MRR 1.0 under the same reference-first ties. The self-distance result therefore checks a construction property and is excluded as evidence of independent discrimination or generation quality.

The final input dimensions are refund-request lifecycle, requested service action, identity-verification completeness, return-logistics progress, problem specificity, sentiment and urgency, product fault versus order mismatch, supporting-evidence availability, product-attribute evidence type, and intent-evidence observability. The output dimensions are identity and order verification, policy alignment, understanding of customer need, product-condition/resolution fit, next-step progression, information completeness, timeline calibration, empathy, clarity and specificity, and contextual consistency.

Every dimension has an operational definition, five anchors, and an evidence specification in the stored space. These anchors describe different semantic scales rather than interchangeable quantities. For example, identity-verification completeness records the visible verification state, while sentiment and urgency record expressed intensity. The output score controls an instruction priority rather than licensing facts absent from the dialogue.

\section{Compilation and Reproduction Details}\label{app:implementation}
The tree is fitted to the clipped predictions of the standardized ridge model on the 60 training examples. Leaf values are compiled into four instruction strengths: scores in $[0,1)$ omit an attribute, $[1,2)$ request a brief acknowledgment, $[2,3)$ request clear attention, and $[3,4]$ request primary emphasis. The original compiler's generic wording is retained in the experiment records; the factual-guard versions qualify its execution using available evidence. The seven rules use splits involving supporting-evidence availability, identity-verification completeness, refund lifecycle, and requested service action.

The language-model identifier is \texttt{doubao-seed-2-1-\allowbreak pro-260915}. The source data are ABCD v1.1. The reproducibility materials specify the dataset revision, deterministic sample selection, split membership, and checksums, together with the full space definitions, role-specific prompts, per-example attributes, fitted mappings, compiled rules, generated responses, and evaluations.

\FloatBarrier
\begingroup\small\raggedright

\endgroup
\end{document}